\documentclass[lettersize,preprint,journal]{IEEEtran}
\IEEEoverridecommandlockouts
\usepackage{amsmath,amsfonts}
\usepackage{algorithmic}
\usepackage{algorithm}
\usepackage{array}
\usepackage[caption=false,font=normalsize,labelfont=sf,textfont=sf]{subfig}
\usepackage{textcomp}
\usepackage{stfloats}
\usepackage{url}
\usepackage{verbatim}
\usepackage{graphicx}
\usepackage{cite}
\usepackage{mhchem}
\usepackage{multirow}
\usepackage[dvipsnames]{xcolor}
\usepackage{lipsum} 
\usepackage{hyperref}
\usepackage{booktabs}
\usepackage{xspace}

\makeatletter
\DeclareRobustCommand\onedot{\futurelet\@let@token\@onedot}
\def\@onedot{\ifx\@let@token.\else.\null\fi\xspace}

\def\eg{\emph{e.g}\onedot} 
\def\ie{\emph{i.e}\onedot} 
 
\def\etc{\emph{etc}\onedot} 
 
\def\etal{\emph{et al}\onedot}
\makeatother

\newcommand{\mysubsubsection}[1]{\noindent {\bf #1}:}

\begin{document}

\title{Glass Segmentation with RGB-Thermal Image Pairs}

\author{Dong Huo, Jian Wang, Yiming Qian and Yee-Hong Yang,~\IEEEmembership{Senior Member,~IEEE}
\thanks{This work was supported in part by the Natural Sciences and Engineering Research Council of Canada, the University of Alberta, and the University of Manitoba. The associate editor coordinating the review of this manuscript and approving it for publication was Dr. Diego Valsesia. (Dong Huo and Jian Wang contributed equally to this work.) (Corresponding author: Dong Huo.)

D. Huo, and Y. Yang are with the Department of Computing Science, University of Alberta, Edmonton, AB T6G 2R3, Canada (e-mail:
dhuo@ualberta.ca; herberty@ualberta.ca).

Jian Wang is with Snapchat NYC, New York, NY 10036, USA (e-mail: jwang4@snapchat.com).

Yiming Qian is with the Department of Computer Science, University of
Manitoba, Winnipeg, MB R3T 2N2, Canada (e-mail: yiming.qian@umanitoba.ca).

This article has supplementary downloadable material available at https://doi.org/10.1109/TIP.2023.3256762, provided by the authors.

Digital Object Identifier 10.1109/TIP.2023.3256762}
}

\markboth{IEEE TRANSACTIONS ON IMAGE PROCESSING}%
{Shell \MakeLowercase{\textit{et al.}}: A Sample Article Using IEEEtran.cls for IEEE Journals}


\maketitle

\begin{abstract}
This paper proposes a new glass segmentation method utilizing paired RGB and thermal images. Due to the large difference between the transmission property of visible light and that of the thermal energy through the glass where most glass is transparent to the visible light but opaque to thermal energy, glass regions of a scene are made more distinguishable with a pair of RGB and thermal images than solely with an RGB image. To exploit such a unique property, we propose a neural network architecture that effectively combines an RGB-thermal image pair with a new multi-modal fusion module based on attention, and integrate CNN and transformer to extract local features and non-local dependencies, respectively. As well, we have collected a new dataset containing 5551 RGB-thermal image pairs with ground-truth segmentation annotations. The qualitative and quantitative evaluations demonstrate the effectiveness of the proposed approach on fusing RGB and thermal data for glass segmentation. Our code and data are available at \href{https://github.com/Dong-Huo/RGB-T-Glass-Segmentation}{https://github.com/Dong-Huo/RGB-T-Glass-Segmentation}.
\end{abstract}

\section{Introduction}\label{sec:intro}



\IEEEPARstart{H}{uman-made} environments are full of architectural elements constructed from glass materials such as glass windows, doors, railings, and walls. Accurately identifying and distinguishing these objects has numerous applications in robotics \cite{sajjan2020clear}, manufacturing \cite{martinez2013industrial} and assistive care \cite{zhang2021trans4trans}. Compared to opaque materials, transparent glasses do not have their own colors and their appearances are acquired from the background, posing inconsistent visual features if the background or viewpoint is changed. 
Therefore, glass objects with background-dependent appearances often pose challenges for visual recognition methods that are tailored to opaque objects.

With the emergence of deep neural networks, recent data-driven methods are capable of segmenting glass regions from a single RGB image and have utilized contextual information~\cite{mei2020don}, reflection detection~\cite{lin2021rich} and boundary supervision~\cite{He_2021_ICCV}. While neural networks are powerful, they are based on  unreliable RGB colors or directly adopt the learning frameworks for opaque materials, resulting in limited accuracy enhancement. Several methods seek to leverage alternative cues such as depth~\cite{wang2013glass}, light-field~\cite{xu2015transcut} or polarized light~\cite{kalra2020deep}. However, these methods are not robust enough for clear glass recognition, which could lead to a noisy segmentation mask (holes, rough boundaries) and a high  false positive rate. 

\begin{figure}[t]
\centering
\includegraphics[width=0.48\textwidth]{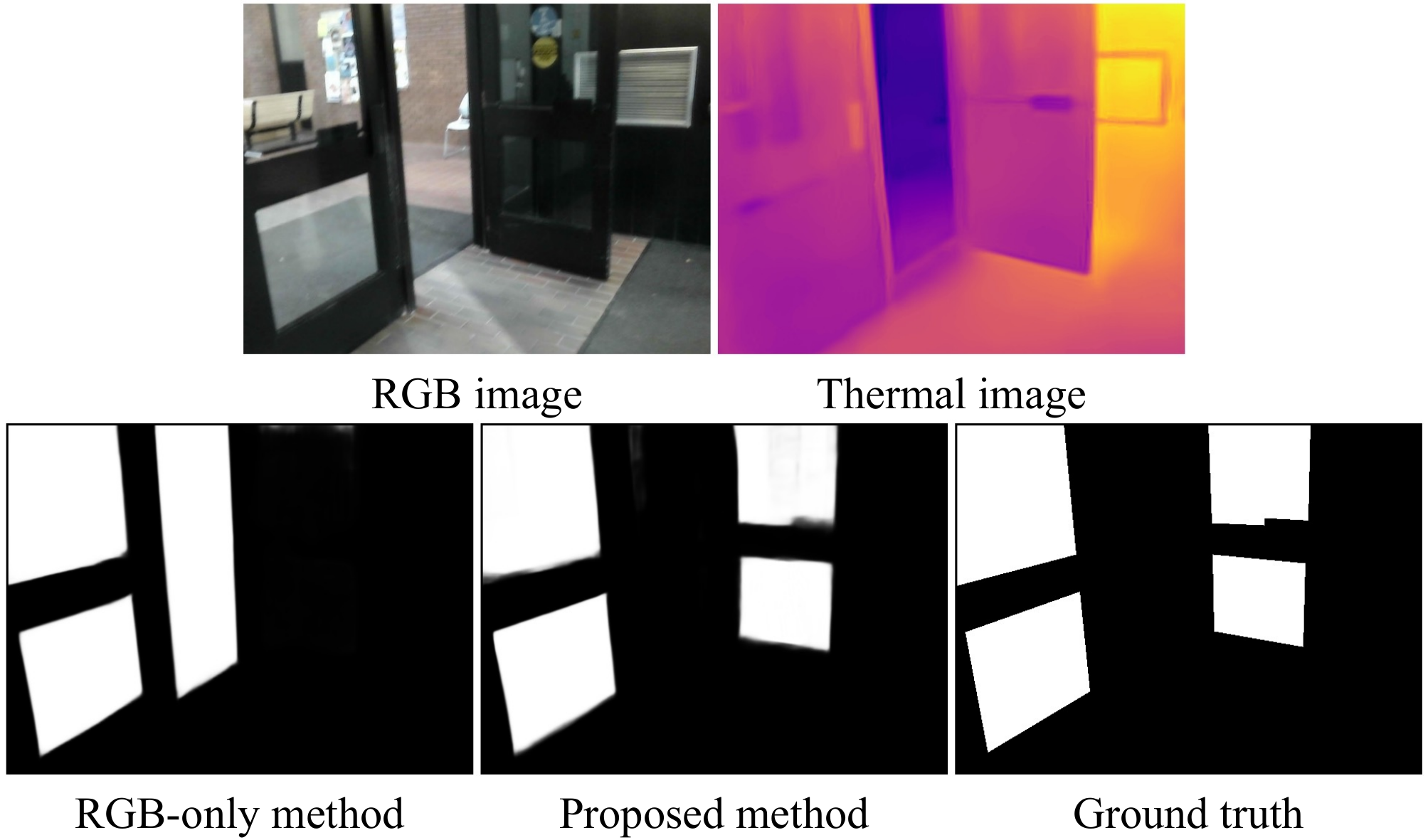}
\caption{The paper proposes a robust glass segmentation approach by fusing a pair of RGB and thermal images, which significantly outperforms the method that takes a single RGB image only.}
\label{fig:teaser}
\end{figure}


This paper takes a step towards fusing RGB and thermal (RGB-T) images for glass segmentation, which is the core contribution. Compared to visible light, which has 100\% transmission, thermal radiation which has the wavelength in the range from 8 to 12$\mu m$ cannot pass through the glass, \ie 0\% transmission. 
Such a unique physical difference between visible light and thermal energy makes glass easier to be detected if an RGB image and a thermal image are jointly processed instead of using the RGB image only. As expected, our experimental results validate our intuition that the proposed RGB-T fusion method outperforms the RGB-only solution by a large margin.

Concretely, we propose a neural network architecture that takes as input a pair of RGB and thermal images (in short RGB-T) and predicts a binary segmentation mask for the glass regions. Following the encoder-decoder framework, our architecture employs (1) two ResNet encoders for feature extraction of the two images, (2) a novel transformer-based fusion module that uses self-attention for correlating the two images at the feature level, and (3) a decoder that uses convolution-based spatial attention for adaptively selecting features for the final mask generation.

We have collected a new dataset consisting of 5551 aligned RGB-T image pairs captured by an off-the-shelf RGB-T camera, where the ground-truth (GT) segmentation masks are obtained by manual annotation. Our dataset is diverse and contains 40 categories of scenes taken from shopping malls, university campuses, residential houses, apartments, public libraries, streets, \etc. The quantitative and qualitative evaluation results demonstrate the effectiveness of using RGB-T image pairs on glass segmentation. Our neural network architecture also excels in multi-modal fusion compared to existing fusion solutions. We also showcase that our approach can greatly improve existing solutions to research problems beyond glass segmentation (\eg, monocular 3D reconstruction and semantic segmentation) when simply applying it as a pre-processing or post-processing step in existing pipelines.

\section{Related Work}

In this section, we discuss related techniques in three  areas: transparent object/glass recognition, RGB-T fusion applications and salient object segmentation.

\subsection{Transparent Object/Glass Recognition}

\mysubsubsection{Methods using RGB only} Traditional algorithms detect glass by analyzing local edge/region characteristics, which exhibit issues in the wild~\cite{fritz2009additive, mchenry2006geodesic, mchenry2005finding, phillips2011novel}. Since the recent progress of deep learning in computer vision, researchers start to collect large-scale transparent object datasets~\cite{xie2020segmenting,mei2020don,lin2021rich,xie2021segmenting} and train their proposed neural networks for glass-like object detection and segmentation from a single RGB image, where contextual feature~\cite{mei2020don,zheng2021glassnet} and boundary supervision~\cite{xie2020segmenting,He_2021_ICCV,zhu2021transfusion} are both popular ways for boosting accuracy. Multi-task learning also has been adopted for transparent object detection and segmentation, where object segmentation is jointly tackled with other related problems such as refractive flow estimation~\cite{chen2018tom}, reflection detection~\cite{lin2021rich} and scene understanding~\cite{zhang2021trans4trans}.


\mysubsubsection{Methods using modalities beyond RGB} The seemingly simple visual appearance of the glass is deceptive because its appearance interacts with the background, which motivates the use of more reliable sensors other than RGB cameras. Existing methods have utilized the refractive distortions captured by a light-field camera~\cite{xu2015transcut} and the high contrast of edges in a polarization image~\cite{kalra2020deep} for transparent object segmentation.  
RGB-Depth (RGB-D) fusion also has been used in both traditional optimization methods~\cite{wang2013glass} and recent learning-based methods~\cite{seib2017friend,sajjan2020clear}. However, depth cameras suffer from severe sensor failures for transparent surfaces due to light refraction. A backlight with AprilTag is employed to enhance RGB-D 3D scanning~\cite{whelan2018reconstructing}. This paper adopts a similar spirit of sensor fusion using a low-cost but robust solution by integrating RGB and thermal images.

\subsection{Salient Object Detection}
Salient object detection (SOD) aims to segment the most prominent object in a given scene. Early methods~\cite{li2014secrets, perazzi2012saliency, yan2013hierarchical} heavily rely on hand-crafted features from a single RGB image. Recent data-driven methods~\cite{qin2019basnet, pang2020multi, zhou2020interactive, tang2021disentangled, siris2021scene} have  dominated this field. 
Integrating RGB and depth images has significantly improved the performance of SOD methods. Among them, direct concatenation~\cite{fu2020jl, zhang2020uc, zhao2020single, chen2021rgb}, addition~\cite{pang2020hierarchical}, spatial/channel attention~\cite{piao2020a2dele, zhang2020select, fan2020bbs, sun2021deep, ji2021calibrated}, prediction guidance~\cite{chen2020progressively}, affine transformation~\cite{li2020rgb, zhou2021specificity}, message passing~\cite{luo2020cascade}, mutual information minimization~\cite{zhang2021rgb}, and self/cross attention~\cite{liu2020learning, zhang2020asymmetric,liu2021visual} have been utilized for feature fusion. 

In addition to depth images, thermal images are also exploited to compensate RGB images for SOD. Tu~\etal~\cite{tu2019rgb} regard deep features of superpixels as graph nodes to cluster the foreground and background with collaborative graph learning. Zhang~\etal~\cite{zhang2020revisiting} learn to generate spatial attention mask for modality fusion of multi-scale features. Zhou~\etal~\cite{zhou2021ecffnet} utilize different dilation rates to extract features from two modalities and combine spatial and channel attention for modality fusion. Zhou~\etal~\cite{zhou2021apnet} adopt a three-branch architecture to generate salient masks from RGB, thermal and fusion features separately and use the weighted summation of three masks as the final results. Sun~\etal~\cite{sun2022hierarchical} utilize sine-cosine functions to extract features from two modalities. This paper focuses on extracting binary masks for glass to tackle the same task of binary segmentation as in SOD. Wu~\etal~\cite{wu2023mfenet} exploit channel attention weights learned from one modality to enhance the other which can better complement features from two modalities. Tu~\etal~\cite{tu2022weakly} propose a modality alignment module for weakly alignment-free RGB-T image pairs.


\subsection{RGB-T Fusion Applications}
RGB-T image pairs are widely used in many vision tasks to compensate the low-quality of RGB images under poor illumination or occlusion, such as object tracking~\cite{li2019rgb, zhang2019multi, wang2020cross, zhang2021jointly, zhang2020rgb, tu2022rgbt, zhang2020dsiammft}, moving object detection~\cite{li2016weighted, zhao2013infrared, yang2017fast, alldieck2016context}, face recognition~\cite{gundimada2010face}, semantic segmentation~\cite{zhou2021gmnet, zhang2021abmdrnet, ha2017mfnet, shivakumar2020pst900, sun2019rtfnet, kutuk2022semantic, sun2020fuseseg}, scene understanding~\cite{gong2023global, zhou2022edge, zhou2021mffenet, zhou2022mtanet, hou2022iaffnet}, crowd counting~\cite{zhou2022defnet, liu2021cross, pahwa2022conditional, liu2022dilated} and salient object detection~\cite{zhang2019rgb, wang2018rgb, tu2019m3s, gao2021unified, guo2021tsfnet, huo2021efficient, wang2021cgfnet, tu2021multi}. To our best knowledge, our proposed method is the first RGB-T method for glass segmentation. As well, we also collected the first dataset for such an application. 
\section{Glass Definition, Scope and Why RGB-T?}
\label{sec:model}

Glass could have different meanings, broad or narrow, and different type of glass has different properties and applications. In this section, we first explain the different glass types, their applications and optical properties. Then we narrow down the glass type to daily transparent glass as the focus of this paper which is also the most difficult to segment. Lastly, we introduce our RGB-T fusion idea and discuss the limitations of other alternative solutions.


\begin{figure}[t]
\centering
\includegraphics[width=0.48\textwidth]{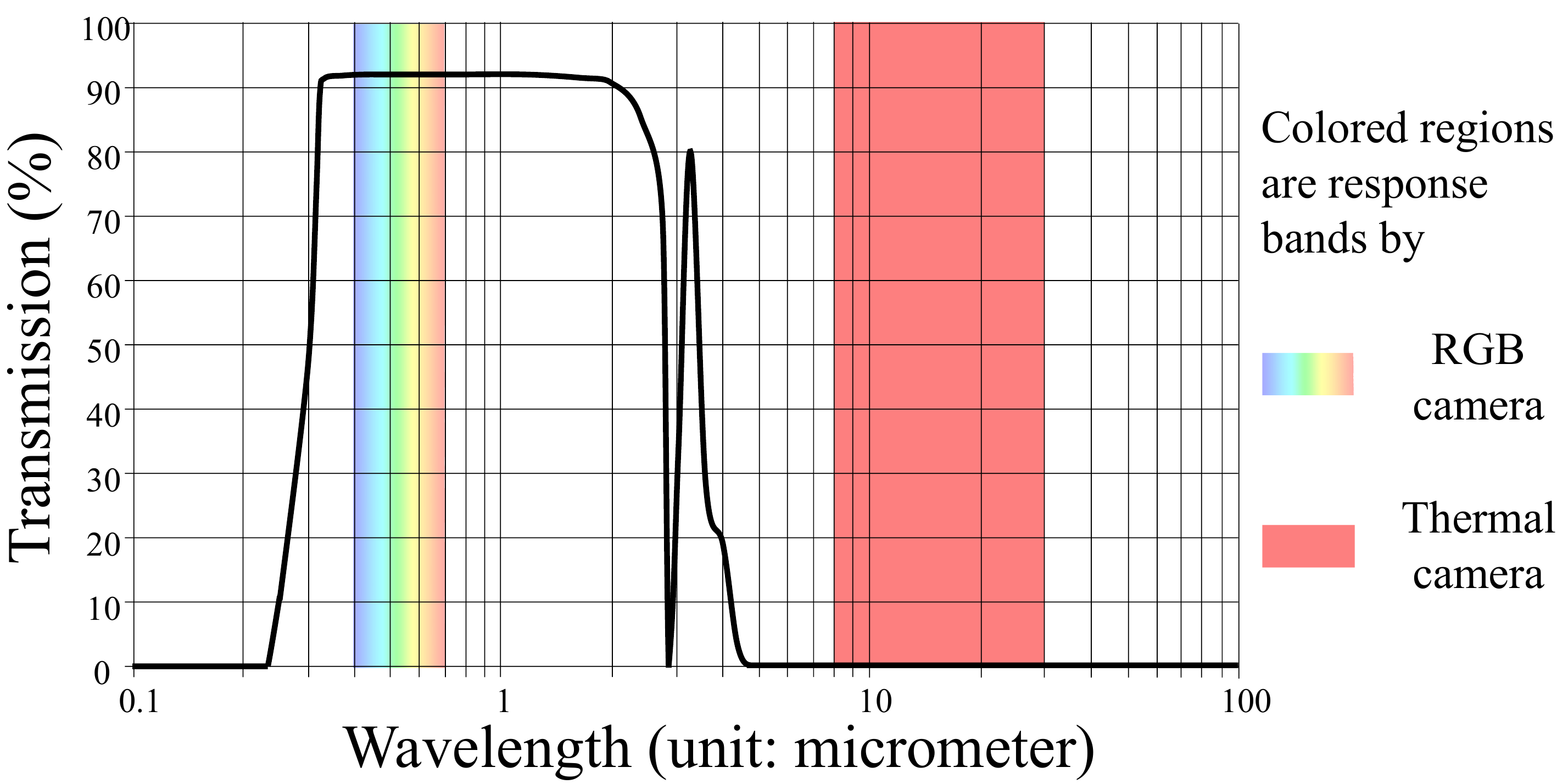}
\caption{
Typical glass spectral transmission curve \cite{vollmer2017infrared} and response bands of RGB and thermal cameras (colored regions).
} 
\label{fig:transmission}
\end{figure}

\begin{figure}[t]
\centering
    \includegraphics[width=0.48\textwidth]{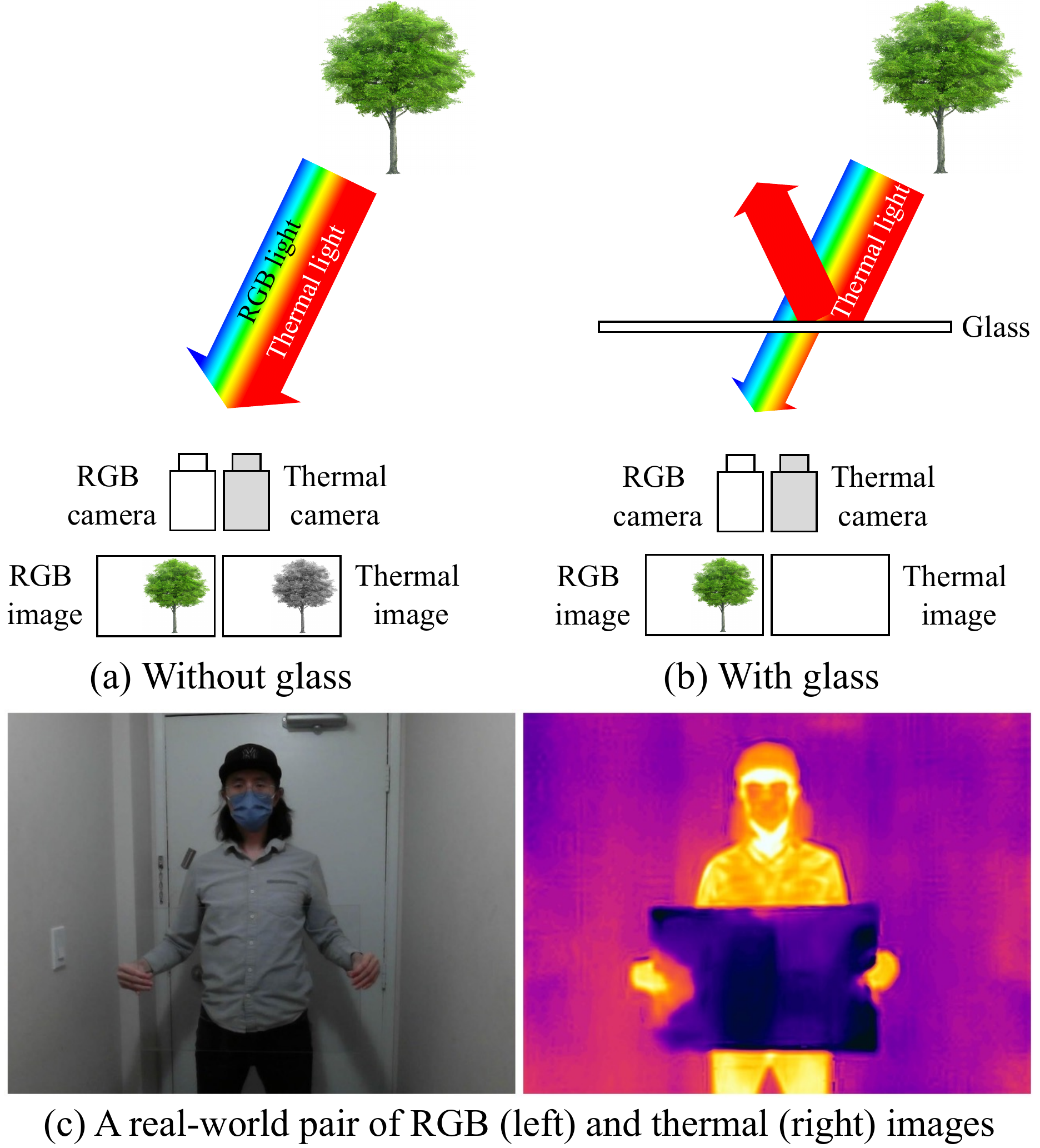}
\caption{A toy illustration for the imaging models of RGB and thermal cameras without and with glass in the scene (a,b). The glass plate held by the person is invisible in the RGB image while visible in the thermal image (c).}
\label{fig:principle}
\end{figure}

Glass is non-crystalline and can be categorized into two types based on its compositions \cite{glassEncyclopedia}. (1) Silicate glass, based on silicon dioxide (\ce{SiO2}) which is abundant on earth in the form of quartz and beach sands. \ce{SiO2} has a very high melting temperature ($\sim 1700^\circ C$), which is hard to work with, and hence, other substances are often added, \eg \ce{Na2O}, \ce{CaO}, \etc, to lower the melting temperature and to tailor to different applications: vitreous silica (100\% \ce{SiO2}) is used for furnace tubes, soda-lime silicate (72\% \ce{SiO2} with \ce{Na2O} and \ce{CaO}) is used for windows, containers and tableware, ``crown'' (69\% \ce{SiO2} with \ce{Na2O}, \ce{B2O3} and \ce{K2O}) is used for optical lens, and aluminosilicate (53\% \ce{SiO2} with \ce{Al2O3}) is used for fibreglass, \etc. (2) Non-silicate based glass includes amorphous metals, chalcogenides, fluorides, polymers, \etc. 

For optical property, glass is transparent to visible light because there are no grain boundaries (the interface between two crystals formed during cooling) which scatter light in polycrystalline materials \cite{carter2007ceramic}. Meanwhile, silicate glasses absorb light with wavelengths longer than 4$\mu m$,  which makes them opaque to long infrared light\footnote{In this paper, these words are interchangeable, ``thermal radiation, thermal energy, long infrared light, long-wave infrared light", all referring to electromagnetic radiation of 8 to 12 $\mu m$.} (Fig. \ref{fig:transmission}). Non-silicate glass could sometimes transmit long infrared light, \eg, fluoride glasses and chalcogenide glasses can transmit light up to 7$\mu m$ and 18$\mu m$, respectively, but they are used for infrared imaging, infrared fiber, and CD/DVDs. 

Some processing can change the transparency slightly. Glass \emph{equally} transmits visible light at different wavelengths (Fig. \ref{fig:transmission}), but can appear tinted after adding some metallic oxides that absorb light of certain wavelengths, \eg, blue by cobalt, green by chromium. Sandblasting or acid etching clear glass creates frosted glass which is translucent. Low-emissivity (Low-e) glass is glass with a thin coating layer to prevent transmission of light 
over 780$nm$ and is usually used as a window to the outside. Note that no coating can increase the transmission rate.


In this paper, {\bf we limit the scope to daily transparent glass}, \ie, the plate glass normally seen in our daily life such as glass windows, doors and tables, which is the most difficult to detect and segment. The daily glass in human-made environments is mostly silicate-based, more specifically, soda-lime silicate (\ce{SiO2} + \ce{Na2O} + \ce{CaO}). Silica glass is like a band-pass optical filter which has a cut-on wavelength 350$nm$ and a cut-off wavelength 4$\mu m$ (Fig. \ref{fig:transmission}). It has high transmission in the visible band ($0.4\mu m\sim 0.7\mu m$) but low in the thermal band ($8\mu m~\sim 12\mu m$). 

Motivated by this fact, {\bf we propose to use an RGB-T image pair to detect and segment glass}. A toy example is shown in Fig.~\ref{fig:principle}(a, b). The tree behind the glass is visible to the RGB camera but not to the thermal camera, which is not the case when the glass is removed. As a result, images captured by thermal cameras have less arbitrary textures than RGB cameras on glass regions, while keeping similar textures on non-glass regions. 
Fig. \ref{fig:principle}(c) shows a real example.

RGB-T is better than the alternative methods. The problem of RGB-UV is that there is limited UV light indoor. The problem of RGB-MWIR (middle-wave IR) is that the MWIR camera is very costly (\$$10^{-1}$ per pixel \cite{gehm2015compressive}). RGB-Near IR (NIR) only works for some Low-e glass. In contrast, RGB-T works for any silicate glass, works indoors and is low cost ($<\$500$), and it is extensible to traditional glass alternatives. We do not know whether some glass in our collected data is made of polymers like plastics and acrylic because we could not check, but acrylic and plastics of several millimeters thick are also opaque to thermal radiation. In the future, there could be new technologies to make glass windows and glass doors; RGB-T should still work because the new materials are expected to have high visibility and low heat transmission, \ie opaque to heat to keep heat out in summer and to prevent heat loss in winter.


\section{RGB-T Glass Segmentation Dataset}

\begin{figure*}
\centering
\includegraphics[width=\textwidth]{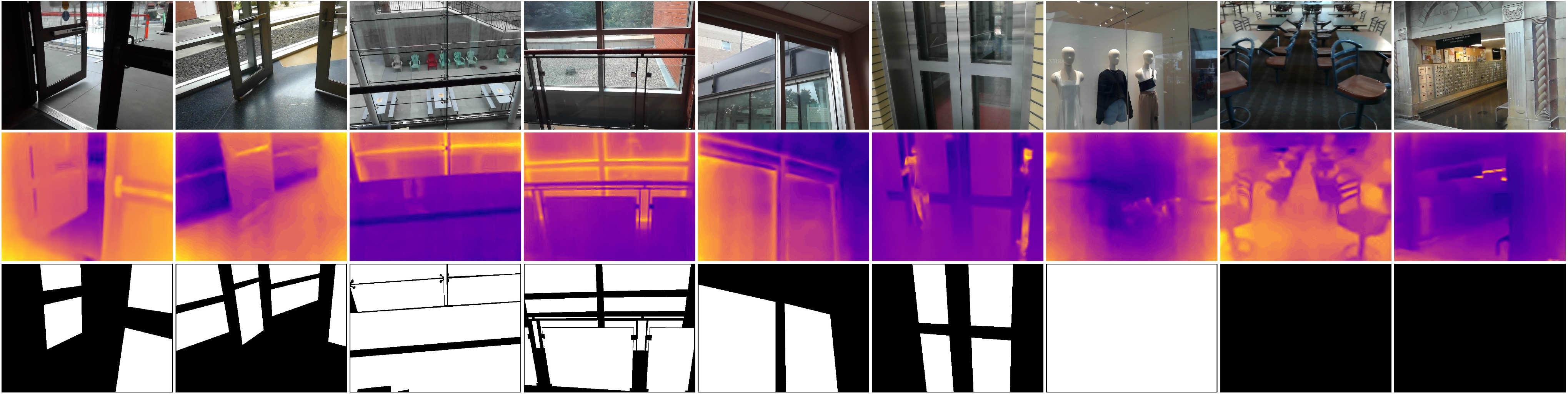}
\caption{Examples of  RGB-T image pairs with GT masks (bottom row) in our dataset. 
The last three columns show images with glass at all pixels or without glass. 
Please note that the image border of each mask is set to black for better visualization.}
\label{fig:data_example}
\end{figure*}

\begin{figure*}
	\centering
	\includegraphics[width=\textwidth]{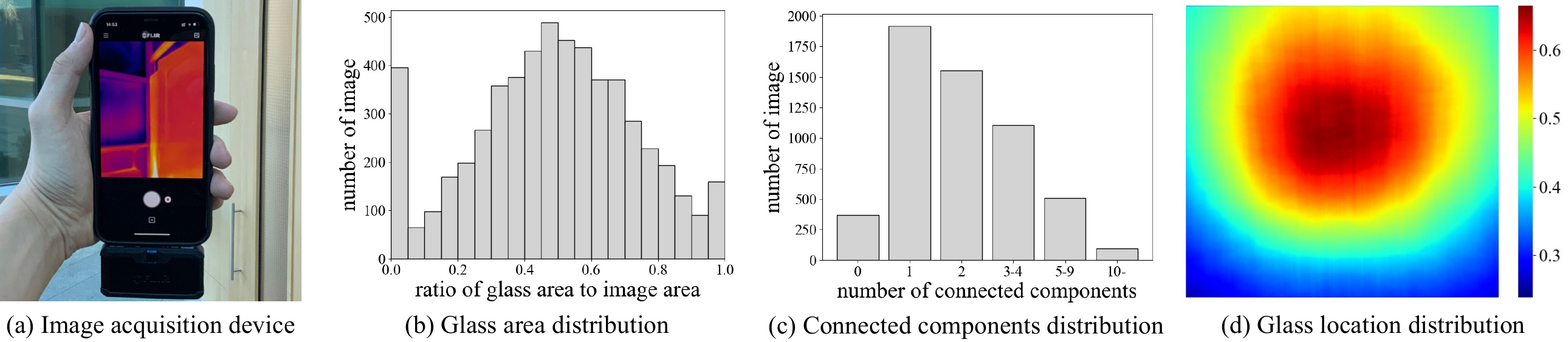}
	\caption{The RGB-T image acquisition device (a) and statistical analysis of our dataset (b,c,d). See text for details.}
    \label{fig:stat}
\end{figure*}

Coupled RGB-T image pairs is a new idea for robust glass segmentation, for which this paper contributes a new RGB-T dataset with GT segmentation masks (see Fig. \ref{fig:data_example}). 

\vspace{1pt}
\mysubsubsection{Dataset construction} We capture RGB-T pairs with a FLIR ONE Pro camera~\cite{flir}, which consists of a nearly collocated RGB sensor and a thermal sensor. The thermal and RGB images are aligned with the FLIR Thermal Studio software~\cite{studio} and both are saved at a resolution of $640\times480$\footnote{The raw thermal image resolution is $160\times120$. The camera software performs super resolution for the thermal image and resizes it to $640\times480$.}. During capturing, the camera is connected to an iPhoneXR for real-time image display (See Fig \ref{fig:stat}(a)). We use LabelMe~\cite{russell2008labelme} to manually annotate segmentation masks over RGB images. Pixels of raw thermal images (only one channel) represent the absolute temperature in the range $-20^\circ C\sim 120^\circ C$, we calculate the min and max values of each thermal image and normalize it to $[0, 1]$ using min-max normalization for obtaining the relative temperature before inputting to the algorithm, which is less sensitive to the variation of ambient temperature. Pseudocolor of thermal images shown in the paper is rendered from grayscale using Matplotlib~\cite{tosi2009matplotlib}.

Our new dataset covers a variety of scenes, such as libraries, shopping malls, galleries, train stations, museums, streets and houses, yielding 5551 RGB-T image pairs from 40 scenes. Among them, we capture 370 pairs without glass in 7 scenes. To generate the train and test split, we randomly select 23 scenes with glass and 5 scenes without glass for training and the others are used for testing (4427/1124 RGB-T pairs for train/test). Note that for the test split, we manually select some images with structures that visually look like glass, \eg, door openings, holes in a wall. Such challenging examples increase the segmentation difficulty for RGB-only solutions and better reflect the merit of our RGB-T fusion idea (\eg, door openings have very different appearances with glass in thermal images).

\mysubsubsection{Dataset statistics} Following~\cite{mei2020don,lin2021rich}, below we conduct some statistical analyses using GT segmentation masks from our dataset:

\noindent \emph{Glass area distribution.} For each mask, we calculate the ratio of the glass area to the entire image area, where a ratio of 0 and 1 indicate, respectively, images without glass and images with glass
at all pixels. As shown in Fig.~\ref{fig:stat}(b), most of the captured images have ratios in the range of $[0.2,0.8]$. There are also quite a few images with a ratio of 0 or 1, which are extreme cases that we captured on purpose.

\noindent \emph{Connected components distribution.} An image may contain several glass regions. We compute the number of connected components in each binary mask and show the histogram in Fig.~\ref{fig:stat}(c). The majority of images have 0$\sim$4 connected components and 52 at most. Images with more than 10 connected components are usually from complex scenes such as that in a shopping mall.

\noindent \emph{Glass location distribution.} 
Fig.~\ref{fig:stat}(d) shows the probability map  of each pixel to be glass. The center region has the highest probability.

\section{RGB-T Glass Segmentation Network}

\begin{figure*}
\centering
\includegraphics[width=\textwidth]{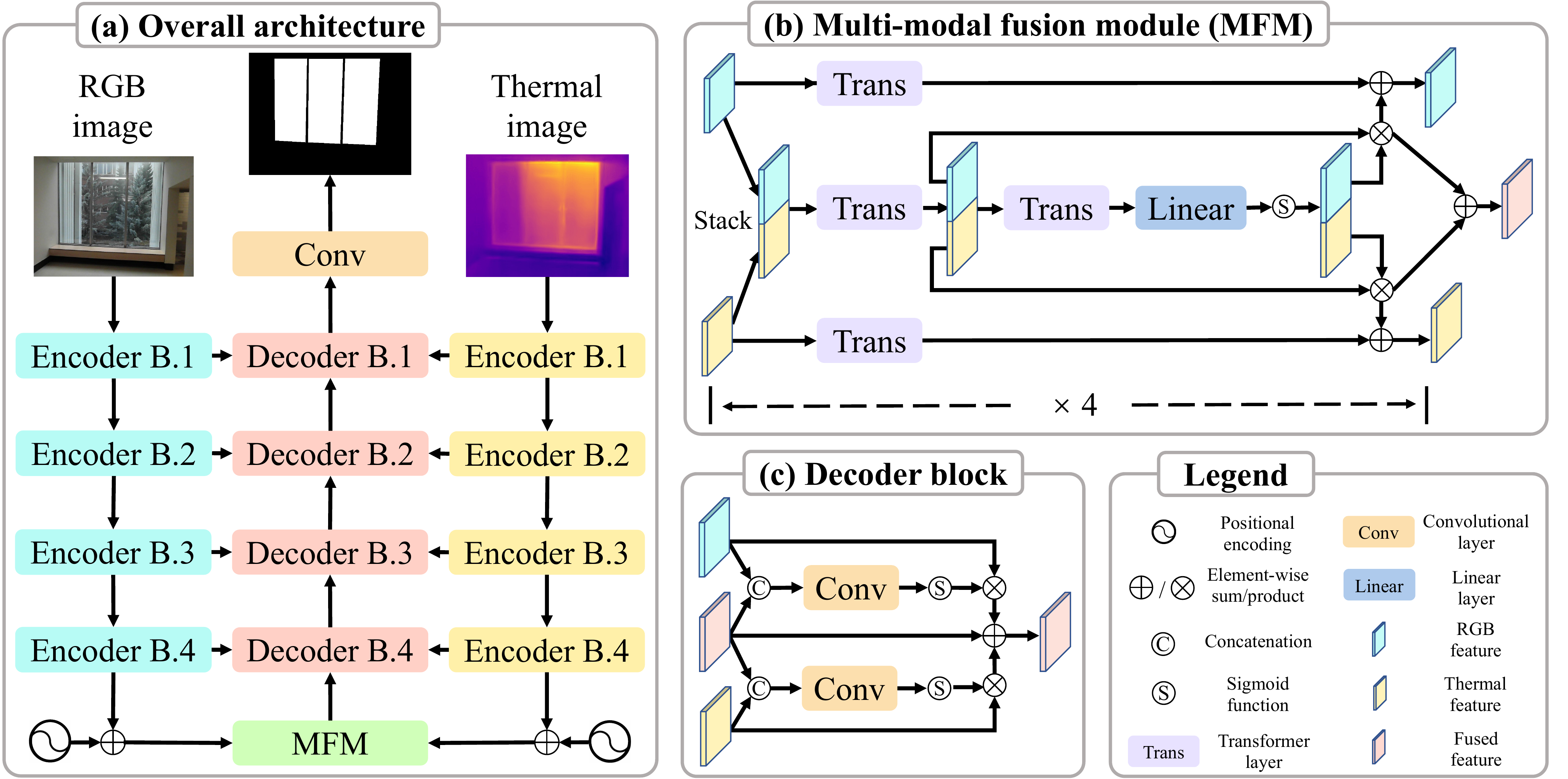}
\caption{Neural network architecture for RGB-T glass segmentation. Our network consists of two separate ResNet-50 backbones as encoders for extracting high-level features from the RGB and thermal images, a transformer-based multi-modal fusion module for integrating the two modalities and a decoder for generating the segmentation result. Encoder/decoder B.$i$ represents the $i$th encoder/decoder block.
}
\label{fig:arch}
\end{figure*}

\subsection{Overall Architecture}
Our architecture follows the standard encoder-decoder framework with skip-connections for dense segmentation~\cite{ronneberger2015u}, which consists of two encoding branches, one decoding branch and a multi-modal fusion module (MFM) as the bridge, as shown in Fig.~\ref{fig:arch}(a). Specifically, we apply two ResNet-50~\cite{he2016deep} encoders to convert the RGB and thermal input images into two $H\times W\times C$ feature volumes, where $H, W$~and $C$, respectively, denote the height, the width and the channel size. In our implementation, $H$ and $W$ are varying and depend on the input image resolution, and $C=256$. We supplement the two features with sinusoidal positional encodings~\cite{parmar2018image} and produce a $H\times W\times C$ fused feature volume by the MFM. The fused feature further passes through a decoding branch with four decoder blocks and is progressively upsampled to the resolution of the input images. Finally, a convolution layer with a sigmoid function converts the results to the final segmentation. Note that all convolution layers other than those 
before the sigmoid functions are followed by a batch normalization layer and a ReLU function, which are omitted in Fig.~\ref{fig:arch}. The Binary Cross-Entropy loss is used for training. Please refer to the supplemental materials for more architectural details.


\subsection{Multi-modal Fusion Module (MFM)}

The foundation of our RGB-T fusion is the attention mechanism in Transformer, which is known to be powerful in combining information from different modalities~\cite{prakash2021multi, liu2021visual}. As shown in Fig.~\ref{fig:arch}(b), our MFM contains four iterative blocks and the $i$th ($i\in\{1,2,3,4\}$) block takes in an RGB feature $\mathbf{f}_r^i$ and a thermal feature $\mathbf{f}_t^i$ both with the same resolution of $HW\times C$\footnote{$\mathbf{f}_r^1$ and $\mathbf{f}_t^1$ are reshaped from the $H\times W\times C$ feature volumes from the two encoders.}. We stack $\mathbf{f}_r^i$ and $\mathbf{f}_t^i$ and use a transformer layer to obtain a $2HW\times C$ feature map $\mathbf{f}_{rt}^i$. We then generate a $2HW\times 1$ weight vector $\mathbf{w}^i$ by passing $\mathbf{f}_{rt}^i$ through a transformer layer followed by a linear layer and a sigmoid function. Such a weight vector is further used to update the RGB and thermal features. Formally, the above process can be summarized as:
\begin{align}
\mathbf{f}_{rt}^i &= \text{trans}(\text{stack}(\mathbf{f}_r^i, \mathbf{f}_t^i)),\\
\mathbf{w}^{i} &= \text{sigmoid}(\text{linear}(\text{trans}(\mathbf{f}_{rt}^{i}))),\\
\mathbf{f}_r^{i+1} &= \text{trans}(\mathbf{f}_r^i) + (\mathbf{w}^{i}\otimes \mathbf{f}_{rt}^{i})[:HW, :],\\
\mathbf{f}_t^{i+1} &= \text{trans}(\mathbf{f}_t^i) + (\mathbf{w}^{i}\otimes \mathbf{f}_{rt}^{i})[HW:, :],
\label{eqn:mfm}
\end{align}
where trans(), linear(), sigmoid(), stack() denote the transformer layer, the linear layer, the sigmoid function and the stack operation, respectively. The symbol $\otimes$ denotes the element-wise multiplication. The transformer layer consists of a multi-head self-attention and a feed-forward network~\cite{vaswani2017attention} (See the supplemental materials for details). Lastly, at the $4$th iteration, the fused feature is computed as $(\mathbf{w}^{4}\otimes \mathbf{f}_{rt}^{4})[:HW, :]+(\mathbf{w}^{4}\otimes \mathbf{f}_{rt}^{4})[HW:, :]$ and reshaped back to $H\times W\times C$.

In our MFM, transformer blocks from the top and the bottom branches extract non-local intra-modality dependencies from the RGB feature $\mathbf{f}_r^i$ and the thermal feature $\mathbf{f}_t^i$, respectively, and we utilize an additional transformer block in the middle branch to extract non-local inter-modality dependencies from the stack of $\mathbf{f}_r^i$ and $\mathbf{f}_t^i$. Considering the discrepancy of two modalities, deficiencies of one modality should be properly compensated by the other. Instead of directly unstacking the multi-modal features $\mathbf{f}_{rt}^i$ along the first dimension and feed them to each modality, we generate a weight vector $\mathbf{w}^{i}$ as the spatial attention mask to remove the detrimental features and keep the beneficial ones. We apply the weight vector (spatial attention mask) $\mathbf{w}^{i}$ on multi-modal features using residual multiplication following Lee~\etal~\cite{lee2020centermask}.


\begin{table*}[t]
\renewcommand\arraystretch{1.5}
\setlength\tabcolsep{5pt}
\centering
\caption{Quantitative evaluations. All compared methods (7 for semantic segmentation, 17 for salient object detection and 1 for glass segmentation) are retrained with our dataset. The performances on images with and without glass are separately evaluated. We also list the results of our thermal-only and RGB-only variants.  The colors \textcolor{blue}{blue} and \textcolor{cyan}{cyan} represent the best and the second best methods, respectively.}
\label{tab:comp}
\begin{tabular}{lccccccccccc} 
\toprule
\multirow{2}{*}{Method} & \multirow{2}{*}{\#Params} & \multirow{2}{*}{\begin{tabular}[c]{@{}c@{}}Inference\\time\end{tabular}} & \multirow{2}{*}{\begin{tabular}[c]{@{}c@{}}Model\\size\end{tabular}} & \multicolumn{4}{c}{Images with glass}                                                                                                                        & \multicolumn{3}{c}{Images without glass}                                                                            & All images        \\ 
\cmidrule[0.25pt](lr){5-8} \cmidrule[0.25pt](lr){9-11} \cmidrule[0.25pt](lr){12-12} 
                        &                           &                                                                          &                                                                      & MAE $\downarrow$                      & IOU $\uparrow$                        & F$_\beta\uparrow$                     & BER $\downarrow$                      & MAE $\downarrow$                      & IOU$^\ast\uparrow$                    & FPR $\downarrow$                     & MAE $\downarrow$  \\ 
\midrule
RTFNet~\cite{sun2019rtfnet}                 & 185.23M                   & 0.016s                                                                   & 708Mb                                                                & 0.068                                 & 88.92                                 & 0.936                                 & 6.675                                 & 0.188                                 & 83.69                                 & 0.89                                 & 0.058             \\ 

ShapeConv~\cite{cao2021shapeconv}              & 41.21M                    & 0.534s                                                                   & 161Mb                                                                & 0.059                                 & 87.65                                 & 0.930                                 & 6.940                                 & 0.019                                 & 98.24                                 & 0.41                                 & 0.054             \\ 

ESANet~\cite{seichter2021efficient}                 & 46.88M                    & 0.022s                                                                   & 359Mb                                                                & 0.051                                 & 90.15                                 & 0.945                                 & 5.863                                 & 0.030                                 & 97.70                                 & 0.25                                 & 0.040             \\ 

CMX~\cite{liu2022cmx}                     & 66.56M                    & 0.035s                                                                   & 255Mb                                                                & \textcolor[rgb]{0,0.675,0.937}{0.031}                                 & \textcolor[rgb]{0,0.675,0.937}{92.40}                                 & \textcolor[rgb]{0,0.675,0.937}{0.956}                                 & 5.421                                 & \textcolor[rgb]{0,0.675,0.937}{0.004}                                 & \textcolor[rgb]{0,0.675,0.937}{99.70}                                 & \textcolor{blue}{0.06}                                 & \textcolor[rgb]{0,0.675,0.937}{0.029}             \\ 
\hline

Segformer~\cite{xie2021segformer}               & 34.07M                    & 0.017s                                                                   & 131Mb                                                                & 0.052                                 & 89.22                                 & 0.934                                & 7.208                                 & 0.063                                 & 93.72                                 & 0.34                                 & 0.053             \\ 

Segmenter~\cite{strudel2021segmenter}               & 103.15M                   & 1.067s                                                                   & 784Mb                                                                & 0.072                                 & 85.83                                 & 0.912                                 & 8.451                                 & 0.303                                 & 92.06                                 & 0.31                                 & 0.072             \\ 
\hline
MCNet~\cite{xiong2021mcnet}                   & 54.64M                    & 0.266s                                                                   & 210Mb                                                                & 0.177                                 & 67.45                                 & 0.782                                 & 19.934                                & 0.118                                 & 89.51                                 & 0.60                                 & 0.172             \\ 
\hline
DPANet~\cite{chen2020dpanet}                 & 92.40M                    & 0.019s                                                                   & 354Mb                                                                & 0.241                                 & 71.51                                 & 0.838                                 & 15.051                                & 0.291                                 & 83.56                                 & 0.90                                 & 0.154             \\ 

HAINet~\cite{li2021hierarchical}                 & 59.82M                    & 0.023s                                                                   & 229Mb                                                                & 0.078                                 & 87.16                                 & 0.932                                 & 7.247                                 & 0.069                                 & 94.44                                 & 0.48                                 & 0.062             \\ 

Zhang~\etal~\cite{zhang2020revisiting}                 & 38.87M                    & 0.053s                                                                   & 445Mb                                                                & 0.156                                 & 75.13                                 & 0.842                                 & 14.141                                & 0.488                                 & 51.57                                 & 1.00                                 & 0.163             \\ 

SSF~\cite{zhang2020select}                   & 32.93M                    & 0.028s                                                                   & 126Mb                                                                & 0.081                                 & 84.53                                 & 0.909                                 & 8.068                                 & 0.356                                 & 64.47                                 & 0.96                                 & 0.097             \\ 

UCNet~\cite{zhang2020uc}                  & 31.26M                    & 0.026s                                                                   & 120Mb                                                                & 0.079                                 & 84.70                                 & 0.913                                 & 8.324                                 & 0.028                                 & 97.24                                 & 0.14                                 & 0.071             \\ 

CoNet~\cite{ji2020accurate}                  & 43.66M                    & 0.021s                                                                   & 168Mb                                                                & 0.118                                 & 79.94                                 & 0.876                                 & 9.336                                 & 0.533                                 & 46.83                                 & 1.00                                 & 0.145             \\ 

ATSA~\cite{zhang2020asymmetric}                   & 32.16M                    & 0.021s                                                                   & 124Mb                                                                & 0.088                                 & 83.87                                 & 0.903                                 & 8.185                                 & 0.276                                 & 73.57                                 & 1.00                                 & 0.098             \\ 

DANet~\cite{zhao2020single}                  & 26.68M                    & 0.007s                                                                   & 102Mb                                                                & 0.082                                 & 85.63                                 & 0.915                                 & 7.982                                 & 0.045                                 & 96.36                                 & 0.36                                 & 0.069             \\ 

HDFNet~\cite{pang2020hierarchical}                & 54.77M                    & 0.019s                                                                   & 419Mb                                                                & 0.055                                 & 89.56                                 & 0.941                                 & 5.673                                 & 0.019                                 & 98.49 & 0.59                                 & 0.048             \\ 

FRDT~\cite{zhang2020feature}                   & 72.81M                    & 0.013s                                                                   & 279Mb                                                                & 0.094                                 & 82.53                                 & 0.890                                 & 9.050                                 & 0.313                                 & 69.49                                 & 0.96                                 & 0.107             \\ 

RD3D~\cite{chen2021rgb}                   & 46.90M                    & 0.013s                                                                   & 180Mb                                                                & 0.064                                 & 88.94                                 & 0.938                                 & 6.610                                 & 0.037                                 & 97.13                                 & 0.26                                 & 0.045             \\ 

DCFNet~\cite{ji2021calibrated}                & 108.49M                   & 0.064s                                                                   & 373Mb                                                                & 0.059                                 & 88.13                                 & 0.930                                 & 6.757                                 & 0.036                                 & 96.45                                 & 0.21                                 & 0.056             \\ 

UTA~\cite{zhao2021rgb}                   & 48.61M                    & 0.023s                                                                   & 98Mb                                                                 & 0.051                                 & 89.59                                 & 0.933                                 & 6.060                                 & 0.069                                 & 93.00                                 & 0.23                                 & 0.052             \\ 

EBS~\cite{zhang2021learning}                     & 118.96M                   & 0.036s                                                                   & 467Mb                                                                & 0.041                                 & 91.24                                 & 0.946                                 & 5.528                                 & 0.031                                 & 96.97                                 & 0.21                                 & 0.040             \\ 

VST~\cite{liu2021visual}                     & 83.83M                    & 0.034s                                                                   & 321Mb                                                                & 0.050                                 & 90.03                                 & 0.939                                 & 5.857                                 & 0.026                                 & 97.47                                 & 0.15                                 & 0.044             \\ 

CLNet~\cite{zhang2021rgb}                  & 246.13M                   & 0.142s                                                                   & 941Mb                                                                & 0.045                                 & 91.01 & 0.945                                 & \textcolor[rgb]{0,0.675,0.937}{4.983} & 0.021                                 & 98.03                                 & 0.14                                 & 0.041             \\ 

SPNet~\cite{zhou2021specificity}                  & 175.29M                   & 0.058s                                                                   & 671Mb                                                                & 0.044 & 90.76                                 & 0.947 & 5.064                                 & 0.035                                 & 96.68                                 & 0.40                                 & 0.041             \\ 

\hline
EBLNet~\cite{He_2021_ICCV}                 & 48.36M                    & 0.012s                                                                   & 185Mb                                                                & 0.113                                 & 80.54                                 & 0.880                                 & 10.301                                & 0.129                                 & 88.67                                 & 0.72                                 & 0.104             \\ 
\hline
Ours (Thermal-only)     & 65.14M                    & 0.039s                                                                   & 238Mb                                                                & 0.189                                 & 68.63                                 & 0.781                                 & 19.315                                & 0.136                                 & 86.68                                 & 0.55                                 & 0.177             \\ 

Ours (RGB-only)         & 65.14M                    & 0.039s                                                                   & 238Mb                                                                & 0.056                                 & 88.94                                 & 0.929                                 & 6.618                                 & 0.016 & 98.42                                 & 0.11 & 0.052             \\ 

Ours (RGB-T)            & 85.02M                    & 0.048s                                                                   & 328Mb                                                                & \textcolor{blue}{0.027}               & \textcolor{blue}{93.80}               & \textcolor{blue}{0.965}               & \textcolor{blue}{4.078}               & \textcolor{blue}{0.003}               & \textcolor{blue}{99.73}               & \textcolor[rgb]{0,0.675,0.937}{0.08}               & \textcolor{blue}{0.024}             \\
\bottomrule
\end{tabular}
\end{table*}

\subsection{Decoder}
Our decoder consists of four blocks, where each block takes in (1) an RGB feature volume $\mathbf{e}_r$ from the RGB encoder, (2) a thermal feature $\mathbf{e}_t$ from the thermal encoder, and (3) a fused feature $\mathbf{d}$ from the previous decoder block. We concatenate $\mathbf{e}_r$ and $\mathbf{d}$ and use three convolution layers and a sigmoid function to compute a single-channel weight volume $\mathbf{w}_r$, which has the same height and width as that of $\mathbf{d}$. We apply the same operation to $\mathbf{e}_t$ and $\mathbf{d}$ and obtain another weight $\mathbf{w}_t$.  Finally, the output of the block is calculated as $\mathbf{w}_r\otimes\mathbf{e}_r+\mathbf{w}_t\otimes\mathbf{e}_t+\mathbf{d}$. Similar to the weight vector $\mathbf{w}^{i}$ in MFM, $\mathbf{w}_r$ and $\mathbf{w}_t$ are used as the spatial attention mask for feature fusion.

Note that we could have chosen transformer layers for feature fusion in the decoder as that in MFM, but we opt for a convolution-based spatial attention scheme due to the intensive memory constraints of self-attention in high resolution~\cite{kitaev2019reformer}. 
In addition, each decoder block is further followed by a convolution layer and a bilinear upsampling to gradually recover the spatial dimension, the detail of which is omitted in Fig.~\ref{fig:arch}(c).


\section{Experiments}
\label{sec:exp}

We have implemented our method in PyTorch~\cite{paszke2019pytorch}. The ImageNet-pretrained ResNet-50 backbone (encoder block 1$\sim$4) is loaded from torchvision. The batch size is 16. The learning rate is initialized as $10^{-4}$ and changed to $10^{-5}$ after 200 epochs. We use random flipping, resizing and cropping for data augmentation. Our model is trained with the  AdamW~\cite{loshchilov2017decoupled} optimizer for 300 epochs, which takes around 35 hours on an NVIDIA RTX A6000 GPU with 48GB of RAM.

\subsection{Evaluation Metrics}

We adopt standard segmentation metrics: mean absolute error (MAE), intersection over union (IOU), maximum F-measure ($F_\beta$) and balanced error rate (BER). These four metrics are commonly used in previous glass segmentation papers~\cite{mei2020don,lin2021rich,He_2021_ICCV}, whereas only MAE is valid for evaluating images without glass (\ie, GT mask is black). To assess our performance on those images without glass in our new dataset, together with MAE, we use inverse intersection over union (IOU$^\ast$) and false positive rate (FPR). IOU$^\ast$ takes inverse masks and is defined as $\text{IOU}(1-\text{result}, 1-\text{GT})$. FPR is calculated as the ratio of the number of false positives (\ie, glass are wrongly detected) to the total number of images without glass. Because  MAE is commonly used, we use it to evaluate on all images.

\begin{figure*}
\centering
\includegraphics[width=\textwidth]{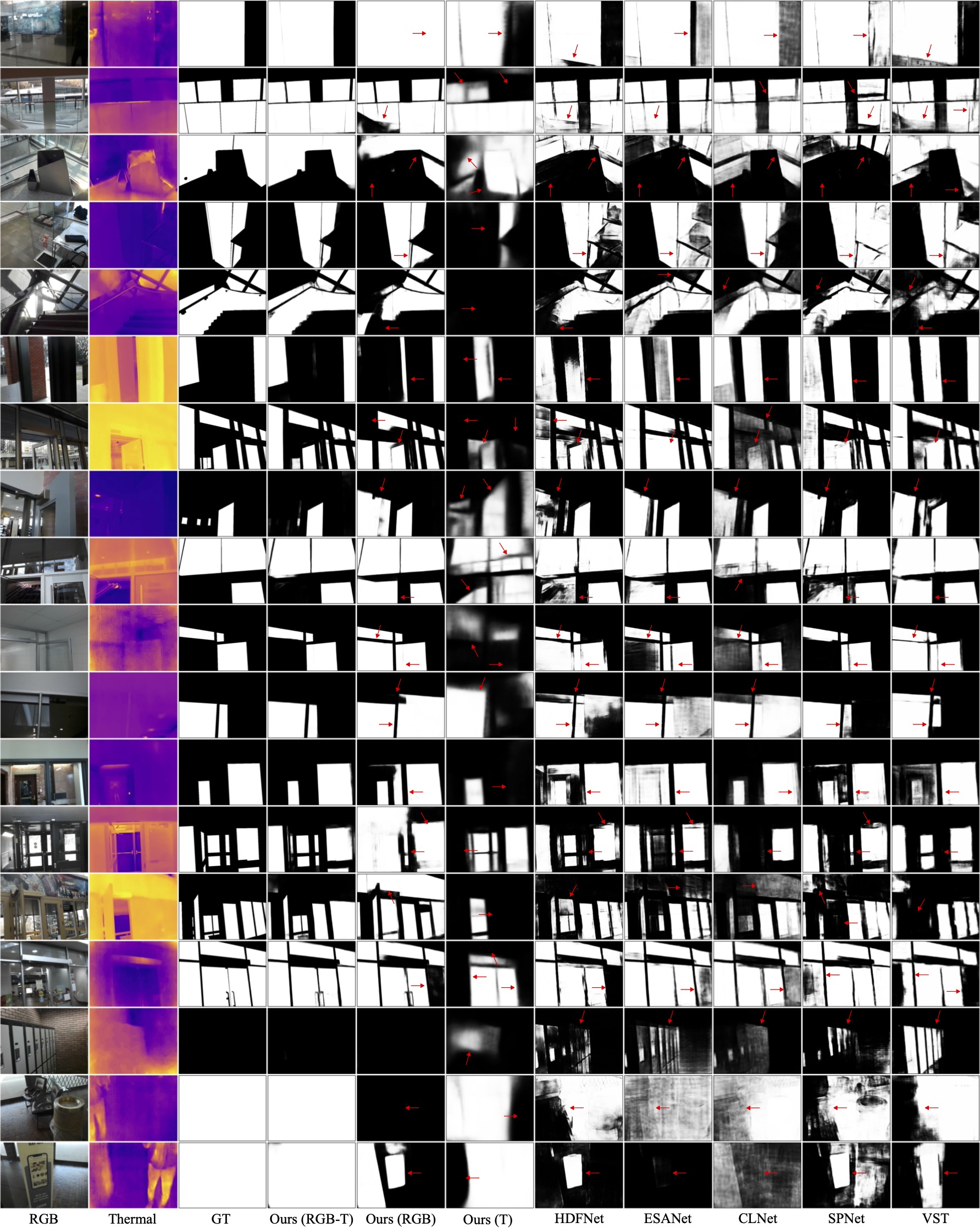}
\caption{Qualitative comparison of our method and 5 state-of-the-art methods (HDFNet~\cite{pang2020hierarchical}, ESANet~\cite{seichter2021efficient}, CLNet~\cite{zhang2021rgb}, SPNet~\cite{zhou2021specificity}, and VST~\cite{liu2021visual}). Results of our RGB-only and thermal-only variants are also displayed. For better visualization, we set the image border of each mask to black. The superiority of our method can be clearly validated at various places, as highlighted by the red arrows.}
\label{fig:qualitative}
\end{figure*}

\begin{table}[t]
\renewcommand\arraystretch{1.5}
\setlength\tabcolsep{5pt}
\centering
\caption{Quantitative evaluations on RGB-only glass segmentation dataset GDD~\cite{mei2020don}. Evaluation results except ours are directly copied from~\cite{He_2021_ICCV}. The colors \textcolor{blue}{blue} and \textcolor{cyan}{cyan} represent the best and the second best methods, respectively.}
\label{tab:GDD}
\begin{tabular}{lccc}
\toprule
Method          & MAE$\downarrow$   & IOU$\uparrow$  & BER$\downarrow$  \\ 
\midrule
MirrorNet~\cite{yang2019my}       & 0.094 & 81.3 & 8.98 \\ 
GDNet~\cite{mei2020don}           & 0.088 & 82.6 & 8.42 \\ 
Translab~\cite{xie2020segmenting}        & 0.081 & 85.1 & 7.43 \\ 
EBLNet~\cite{He_2021_ICCV}          & \textcolor{cyan}{0.074} & \textcolor{cyan}{86.0} & \textcolor{blue}{6.90} \\ \hline
Ours (RGB-only) & \textcolor{blue}{0.072} & \textcolor{blue}{86.6} & \textcolor{cyan}{7.35} \\ 
\bottomrule
\end{tabular}
\end{table}

\subsection{Quantitative Evaluations}
\label{sec:quan}
RGB-T image pairs are new cues for glass segmentation. We compare three variants of our method with 24 state-of-the-art methods from other related areas, which include ShapeConv~\cite{cao2021shapeconv}, ESANet~\cite{seichter2021efficient} and CMX~\cite{liu2022cmx} for RGB-D semantic segmentation, RTFNet~\cite{sun2019rtfnet} for RGB-T semantic segmentation, Segformer~\cite{xie2021segformer} and Segmenter~\cite{strudel2021segmenter} for RGB-only semantic segmentation, MCNet~\cite{xiong2021mcnet} for thermal-only semantic segmentation, DPANet~\cite{chen2020dpanet}, HAINet~\cite{li2021hierarchical}, SSF~\cite{zhang2020select}, UCNet~\cite{zhang2020uc}, CoNet~\cite{ji2020accurate}, ATSA~\cite{zhang2020asymmetric}, DANet~\cite{zhao2020single}, HDFNet~\cite{pang2020hierarchical}, FRDT~\cite{zhang2020feature}, RD3D~\cite{chen2021rgb}, DCFNet~\cite{ji2021calibrated}, UTA~\cite{zhao2021rgb}, EBS~\cite{zhang2021learning}, VST~\cite{liu2021visual}, CLNet~\cite{zhang2021rgb} and SPNet~\cite{zhou2021specificity} for RGB-D salient object detection, and Zhang~\etal~\cite{zhang2020revisiting} for RGB-T saliency detection. All of these competing methods are retrained with our dataset using RGB-T pairs as input. We also compare with a state-of-the-art RGB-only glass segmentation method EBLNet\footnote{Other recent glass segmentation methods~\cite{yang2019my, mei2020don,lin2021rich} did not provide training codes for their models.}~\cite{He_2021_ICCV}, which is retrained with the RGB images in our dataset. 

As shown in Table~\ref{tab:comp}, the last three rows give the ablation results of our full RGB-T method and its two variants that use either thermal or RGB data only. Albeit the thermal-only variant is the worst, it can significantly boost the performance when it is combined with RGB images. For example, the MAE of our RGB-T method on glass images is $86\%$ and $52\%$ better than the two variants, which demonstrates the effectiveness of  our RGB-T fusion idea for glass segmentation. Please refer to the Section~\ref{sec:ablation} for more ablation studies.

Our approach consistently outperforms previous methods on most metrics by a large margin. Similar to our method, CMX, EBS and Segformer utilize a combination of convolution and transformer (hybrid) in their architectures, which achieve superior performance in the evaluations of RGB-T and RGB-only, respectively. We attribute it to the hybrid architecture~\cite{carion2020end, xu2021line}, while pure convolution (\eg, ESANet, CLNet, SPNet, EBLNet) or transformer (\eg, VST, Segmenter) architectures obtain inferior results. 

The performance of our thermal-only variant and MCNet is much worse than that of other RGB-only and RGB-T methods. We believe that there are two reasons. First, the resolution of thermal images is lower than that of RGB images. Hence, the segmentation results of thermal images are poor. Second, even though the glass is opaque to the thermal light, we need to compare it with the RGB images to see the contrast. Hence, it is difficult, if not impossible, to distinguish glass and other opaque objects using a single thermal image.

\begin{table}[t]
\renewcommand\arraystretch{1.5}
\setlength\tabcolsep{5pt}
\centering
\caption{Quantitative evaluations on RGB-T SOD dataset VT5000~\cite{tu2022rgbtbench} and VT1000~\cite{tu2019rgb}.  The colors \textcolor{blue}{blue} and \textcolor{cyan}{cyan} represent the best and the second best methods, respectively.}
\label{tab:rgbt_sod}
\begin{tabular}{lcccccc} 
\toprule
\multirow{2}{*}{Method} & \multicolumn{3}{c}{VT5000~\cite{tu2022rgbtbench}}                           & \multicolumn{3}{c}{VT1000~\cite{tu2019rgb}}                            \\ 
\cmidrule[0.25pt](lr){2-4} \cmidrule[0.25pt](lr){5-7}
                        & MAE $\downarrow$ & S$_m\uparrow$ & F$_\beta\uparrow$ & MAE $\downarrow$ & S$_m\uparrow$ & F$_\beta\uparrow$  \\ 
\midrule
MTMR~\cite{wang2018rgb}                    & 0.114            & 0.680         & 0.662             & 0.119            & 0.706         & 0.755               \\ 

M3S-NIR~\cite{tu2019m3s}                 & 0.188            & 0.631         & 0.607             & 0.151            & 0.717         & 0.734               \\ 

SGDL~\cite{tu2019rgb}                    & 0.089            & 0.750         & 0.738             & 0.090            & 0.787         & 0.807               \\ 

ADF~\cite{tu2022rgbtbench}                     & 0.048            & 0.864         & 0.864             & 0.034            & 0.910         & 0.923               \\ 

MIDD~\cite{tu2021multi}                    & 0.043            & 0.868         & 0.872             & 0.027            & 0.915         & 0.926               \\ 

APNet~\cite{zhou2021apnet}                   & \textcolor{blue}{0.035}            & 0.876         & 0.875             & \textcolor{cyan}{0.021}            & 0.921         & 0.930               \\ 

ECFFNet~\cite{zhou2021ecffnet}                 & 0.038            & 0.874         & 0.872             & \textcolor{cyan}{0.021}            & 0.923         & 0.930               \\ 

MIA-DPD~\cite{liang2022multi}                 & 0.040            & \textcolor{cyan}{0.879}         & \textcolor{cyan}{0.880}             & 0.025            & 0.924         & \textcolor{cyan}{0.935}               \\ 

DCNet~\cite{tu2022weakly}                   & \textcolor{blue}{0.035}            & 0.872         & 0.870             & \textcolor{cyan}{0.021}            & 0.922         & 0.928               \\ 
\hline
Ours                    & \textcolor{cyan}{0.036}            & \textcolor{blue}{0.881}         & \textcolor{blue}{0.881}             & \textcolor{blue}{0.020}            & \textcolor{blue}{0.929}         & \textcolor{blue}{0.941}               \\
\bottomrule
\end{tabular}
\end{table}

\subsection{Qualitative Evaluations}
Figure~\ref{fig:qualitative} shows the qualitative comparison results. Our method is able to accurately segment the glass regions in various challenging scenes, while previous methods and the RGB-only variant produce a plethora of noticeable errors: (1) blurry segmentation results and fuzzy boundaries, (2) under-segmentation masks due to the influence by the background behind glass, and (3) over-segmentation results where cabinet or door openings are wrongly identified as glass.
The last two rows show two extreme cases where the scenes are completely covered by glass. 
The human bodies in the thermal images are reflections of the photographer, which are invisible in the RGB images, further validating the different transmission models of visual and thermal light through glass. Our thermal-only variant tends to recognize the distinct boundaries which may mislead the segmentation. We provide more visual results in the supplementary materials.

\begin{table*}[ht]
\renewcommand\arraystretch{1.5}
\setlength\tabcolsep{5pt}
\centering
\caption{Ablation studies on input. The colors \textcolor{blue}{blue} and \textcolor{cyan}{cyan} represent the best and the second best methods, respectively.}
\label{tab:ab_input}             
\begin{tabular}{lcccccccc} 
\toprule
\multirow{2}{*}{Method} & \multicolumn{4}{c}{Images with glass} & \multicolumn{3}{c}{Images without glass} & All images        \\ 
\cmidrule[0.25pt](lr){2-5} \cmidrule[0.25pt](lr){6-8} \cmidrule[0.25pt](lr){9-9} 
& MAE $\downarrow$                      & IOU $\uparrow$                        & F$_\beta\uparrow$                     & BER $\downarrow$                      & MAE $\downarrow$                      & IOU$^\ast\uparrow$                    & FPR $\downarrow$                     & MAE $\downarrow$  \\ 
\midrule
     
Thermal-only & 0.189 & 68.63 & 0.781 & 19.315 & 0.136 & 86.68 & 0.55 & 0.018\\ 
Dual-Thermal & 0.189 & 68.47 & 0.783 & 19.120 & 0.120 & 88.34 & 0.50 & 0.018\\ 
RGB-only & \multicolumn{1}{c}{0.056} & \multicolumn{1}{c}{88.94} & \multicolumn{1}{c}{0.929} & 6.618 & \multicolumn{1}{c}{\textcolor{cyan}{0.016}} & \multicolumn{1}{c}{\textcolor{cyan}{98.42}} & 0.11 & 0.052\\ 
Dual-RGB & \multicolumn{1}{c}{\textcolor{cyan}{0.055}} & \multicolumn{1}{c}{\textcolor{cyan}{89.21}} & \multicolumn{1}{c}{\textcolor{cyan}{0.932}} & \textcolor{cyan}{6.250} & \multicolumn{1}{c}{0.018} & \multicolumn{1}{c}{98.25} & \textcolor{cyan}{0.10} & \textcolor{cyan}{0.050}\\ \hline

RGB-T (Ours) & \multicolumn{1}{c}{\textcolor{blue}{0.027}} & \multicolumn{1}{c}{\textcolor{blue}{93.80}} & \multicolumn{1}{c}{\textcolor{blue}{0.965}} & \textcolor{blue}{4.078} & \multicolumn{1}{c}{\textcolor{blue}{0.003}} & \multicolumn{1}{c}{\textcolor{blue}{99.73}} & \textcolor{blue}{0.07} & \textcolor{blue}{0.024}\\ 
\bottomrule
\end{tabular}
\end{table*}

\subsection{Evaluations on GDD dataset~\cite{mei2020don}}
\label{sec:GDD}
Table~\ref{tab:GDD} compares our RGB-only variant with four state-of-the-art glass segmentation methods: MirrorNet~\cite{yang2019my}, GDNet~\cite{mei2020don}, Translab~\cite{xie2020segmenting} and EBLNet~\cite{He_2021_ICCV}. Both our variant and compared methods are trained and tested on the RGB-only glass segmentation dataset GDD~\cite{mei2020don}. Our RGB-only variant removes the thermal branch from our RGB-T architecture and the resulted RGB-only architecture is a combination of convolution and transformer, while the four existing methods adopt convolution only. 
Our RGB-only variant achieves the best in IOU and MAE and the second best in BER, echoing recent other recognition methods (\eg, object detection~\cite{carion2020end}, wireframe parsing~\cite{xu2021line}) which also demonstrate the effectiveness of the hybrid structure of convolution and transformer.

\subsection{Evaluations on RGB-T SOD datasets}
To show the versatility of our method on other RGB-T tasks, we retrain our model on the 2500 training images from VT5000 dataset~\cite{tu2022rgbtbench} for RGB-T salient object detection (SOD), then evaluate on the 2500 testing images from VT5000 dataset and 1000 images from VT1000 dataset~\cite{tu2019rgb}. Nine state-of-the-art RGB-T SOD methods are compared including three conventional graph-based methods (MTMR~\cite{wang2018rgb}, M3S-NIR~\cite{tu2019m3s}, SGDL~\cite{tu2019rgb}) and six deep learning based methods (ADF~\cite{tu2022rgbtbench}, MIDD~\cite{tu2021multi}, APNet~\cite{zhou2021apnet}, ECFFNet~\cite{zhou2021ecffnet}, MIA-DPD~\cite{liang2022multi}, DCNet~\cite{tu2022weakly}). We take the mean absolute error (MAE), S-measure ($S_m$)~\cite{fan2017structure}, and maximum F-measure ($F_\beta$) to evaluate the SOD results. As shown in Table~\ref{tab:rgbt_sod}, our method is comparable with others specifically designed for RGB-T SOD.

\begin{table*}[ht]
\renewcommand\arraystretch{1.5}
\setlength\tabcolsep{5pt}
\centering
\caption{Ablation studies on MFM. The colors \textcolor{blue}{blue} and \textcolor{cyan}{cyan} represent the best and the second best methods, respectively.}
\label{tab:ab_mfm}
\begin{tabular}{lcccccccc} 
\toprule
\multirow{2}{*}{Method} & \multicolumn{4}{c}{Images with glass} & \multicolumn{3}{c}{Images without glass} & All images        \\ 
\cmidrule[0.25pt](lr){2-5} \cmidrule[0.25pt](lr){6-8} \cmidrule[0.25pt](lr){9-9} 
& MAE $\downarrow$                      & IOU $\uparrow$                        & F$_\beta\uparrow$                     & BER $\downarrow$                      & MAE $\downarrow$                      & IOU$^\ast\uparrow$                    & FPR $\downarrow$                     & MAE $\downarrow$  \\ 
\midrule
SFS & \multicolumn{1}{c}{0.058} & \multicolumn{1}{c}{89.88} & \multicolumn{1}{c}{0.946} & 5.865 & \multicolumn{1}{c}{0.041} & \multicolumn{1}{c}{96.08} & 0.29 & 0.045\\ 
SFC & \multicolumn{1}{c}{0.052} & \multicolumn{1}{c}{90.47} & \multicolumn{1}{c}{0.945} & 5.588 & \multicolumn{1}{c}{0.025} & \multicolumn{1}{c}{98.15} & 0.36 & 0.043\\ 
PAF~\cite{kalra2020deep} & \multicolumn{1}{c}{0.038} & \multicolumn{1}{c}{91.73} & \multicolumn{1}{c}{0.954} & 5.188 & \multicolumn{1}{c}{0.010} & \multicolumn{1}{c}{99.04} & \textcolor{blue}{0.05} & 0.035\\ 
AT~\cite{li2020rgb} & \multicolumn{1}{c}{0.038} & \multicolumn{1}{c}{91.96} & \multicolumn{1}{c}{0.953} & 4.885 & \multicolumn{1}{c}{0.016} & \multicolumn{1}{c}{98.48} & 0.20 & 0.033\\ 
MFM-EGFNet~\cite{zhou2022edge} & \multicolumn{1}{c}{\textcolor{cyan}{0.030}} & \multicolumn{1}{c}{92.82} & \multicolumn{1}{c}{\textcolor{cyan}{0.960}} & 5.328 & \multicolumn{1}{c}{0.006} & \multicolumn{1}{c}{99.42} & 0.10 & 0.028\\
MFM-DS & \multicolumn{1}{c}{0.031} & \multicolumn{1}{c}{\textcolor{cyan}{93.06}} & \multicolumn{1}{c}{0.959} & \textcolor{cyan}{4.392} & \multicolumn{1}{c}{\textcolor{cyan}{0.004}} & \multicolumn{1}{c}{99.59} & \textcolor{blue}{0.05} & \textcolor{cyan}{0.027}\\ 
MFM-DC & \multicolumn{1}{c}{0.033} & \multicolumn{1}{c}{92.75} & \multicolumn{1}{c}{0.958} & 4.565 & \multicolumn{1}{c}{\textcolor{blue}{0.003}} & \multicolumn{1}{c}{\textcolor{cyan}{99.69}} & \textcolor{cyan}{0.06} & 0.029\\ \hline

MFM (Ours) & \multicolumn{1}{c}{\textcolor{blue}{0.027}} & \multicolumn{1}{c}{\textcolor{blue}{93.80}} & \multicolumn{1}{c}{\textcolor{blue}{0.965}} & \textcolor{blue}{4.078} & \multicolumn{1}{c}{\textcolor{blue}{0.003}} & \multicolumn{1}{c}{\textcolor{blue}{99.73}} & 0.07 &\textcolor{blue}{0.024}\\ 
\bottomrule
\end{tabular}
\end{table*}


\begin{table*}[ht]
\renewcommand\arraystretch{1.5}
\setlength\tabcolsep{5pt}
\centering
\caption{Ablation studies on backbones. The colors \textcolor{blue}{blue} and \textcolor{cyan}{cyan} represent the best and the second best methods, respectively.} 
\label{tab:ab_backbone}
\begin{tabular}{lcccccccc} 
\toprule
\multirow{2}{*}{Method} & \multicolumn{4}{c}{Images with glass} & \multicolumn{3}{c}{Images without glass} & All images        \\ 
\cmidrule[0.25pt](lr){2-5} \cmidrule[0.25pt](lr){6-8} \cmidrule[0.25pt](lr){9-9} 
& MAE $\downarrow$                      & IOU $\uparrow$                        & F$_\beta\uparrow$                     & BER $\downarrow$                      & MAE $\downarrow$                      & IOU$^\ast\uparrow$                    & FPR $\downarrow$                     & MAE $\downarrow$  \\ 
\midrule
ResNet-18  & 0.033 & 92.34 & 0.956 & 5.388 & \textcolor{cyan}{0.006} & \textcolor{cyan}{99.48} & \textcolor{cyan}{0.09} & 0.031\\
ResNet-34 & 0.031 & 92.96 & 0.959 & 5.049 & 0.011 & 98.90 & 0.12 & 0.030\\
ResNet-101 & \multicolumn{1}{c}{\textcolor{cyan}{0.030}} & \multicolumn{1}{c}{\textcolor{cyan}{93.14}} & \multicolumn{1}{c}{\textcolor{cyan}{0.961}} & \textcolor{cyan}{4.982} & \multicolumn{1}{c}{0.010} & \multicolumn{1}{c}{99.04} & \textcolor{blue}{0.07} & \textcolor{cyan}{0.029}\\
ResNet-50 w/o pretraining  & \multicolumn{1}{c}{0.057} & \multicolumn{1}{c}{87.29} & \multicolumn{1}{c}{0.924} & 8.063 & \multicolumn{1}{c}{0.064} & \multicolumn{1}{c}{94.02} & 0.50 & 0.057\\ \hline
ResNet-50 (Ours) & \multicolumn{1}{c}{\textcolor{blue}{0.027}} & \multicolumn{1}{c}{\textcolor{blue}{93.80}} & \multicolumn{1}{c}{\textcolor{blue}{0.965}} & \textcolor{blue}{4.078} & \multicolumn{1}{c}{\textcolor{blue}{0.003}} & \multicolumn{1}{c}{\textcolor{blue}{99.73}} & \textcolor{blue}{0.07} &\textcolor{blue}{0.024}\\
\bottomrule
\end{tabular}
\end{table*}

\begin{table*}[ht]
\renewcommand\arraystretch{1.5}
\setlength\tabcolsep{5pt}
\centering
\caption{Ablation studies on decoder. The colors \textcolor{blue}{blue} and \textcolor{cyan}{cyan} represent the best and the second best methods, respectively.} 
\label{tab:ab_decoder}
\begin{tabular}{lcccccccc} 
\toprule
\multirow{2}{*}{Method} & \multicolumn{4}{c}{Images with glass} & \multicolumn{3}{c}{Images without glass} & All images        \\ 
\cmidrule[0.25pt](lr){2-5} \cmidrule[0.25pt](lr){6-8} \cmidrule[0.25pt](lr){9-9} 
& MAE $\downarrow$                      & IOU $\uparrow$                        & F$_\beta\uparrow$                     & BER $\downarrow$                      & MAE $\downarrow$                      & IOU$^\ast\uparrow$                    & FPR $\downarrow$                     & MAE $\downarrow$  \\ 
\midrule

Decoder-DS & \multicolumn{1}{c}{\textcolor{cyan}{0.033}} & \multicolumn{1}{c}{\textcolor{cyan}{92.87}} & \multicolumn{1}{c}{\textcolor{cyan}{0.960}} & \textcolor{cyan}{4.460} & \multicolumn{1}{c}{\textcolor{cyan}{0.007}} & \multicolumn{1}{c}{\textcolor{cyan}{99.30}} & 0.11 & \textcolor{cyan}{0.030}\\ 
Decoder-DC & \multicolumn{1}{c}{0.034} & \multicolumn{1}{c}{92.42} & \multicolumn{1}{c}{0.956} & 4.907 & \multicolumn{1}{c}{0.008} & \multicolumn{1}{c}{99.24} & \textcolor{cyan}{0.10} & 0.031\\ \hline

Decoder (Ours) & \multicolumn{1}{c}{\textcolor{blue}{0.027}} & \multicolumn{1}{c}{\textcolor{blue}{93.80}} & \multicolumn{1}{c}{\textcolor{blue}{0.965}} & \textcolor{blue}{4.078} & \multicolumn{1}{c}{\textcolor{blue}{0.003}} & \multicolumn{1}{c}{\textcolor{blue}{99.73}} & \textcolor{blue}{0.07} &\textcolor{blue}{0.024}\\ 
\bottomrule
\end{tabular}
\end{table*}

\begin{table*}[ht]
\renewcommand\arraystretch{1.5}
\setlength\tabcolsep{5pt}
\centering
\caption{Ablation studies on images without glass. The colors \textcolor{blue}{blue} and \textcolor{cyan}{cyan} represent the best and the second best methods, respectively.} 
\label{tab:ab_glass}
\begin{tabular}{lcccccccc} 
\toprule
\multirow{2}{*}{Method} & \multicolumn{4}{c}{Images with glass} & \multicolumn{3}{c}{Images without glass} & All images        \\ 
\cmidrule[0.25pt](lr){2-5} \cmidrule[0.25pt](lr){6-8} \cmidrule[0.25pt](lr){9-9} 
& MAE $\downarrow$                      & IOU $\uparrow$                        & F$_\beta\uparrow$                     & BER $\downarrow$                      & MAE $\downarrow$                      & IOU$^\ast\uparrow$                    & FPR $\downarrow$                     & MAE $\downarrow$  \\ 
\midrule

Training w/ glass images only & \multicolumn{1}{c}{\textcolor{cyan}{0.029}} & \multicolumn{1}{c}{\textcolor{cyan}{93.03}} & \multicolumn{1}{c}{\textcolor{cyan}{0.960}} & \textcolor{cyan}{5.116} & \multicolumn{1}{c}{\textcolor{cyan}{0.006}} & \multicolumn{1}{c}{\textcolor{cyan}{99.36}} & \textcolor{cyan}{0.09} & \textcolor{cyan}{0.028}\\ \hline
Training w/ all images (Ours) & \multicolumn{1}{c}{\textcolor{blue}{0.027}} & \multicolumn{1}{c}{\textcolor{blue}{93.80}} & \multicolumn{1}{c}{\textcolor{blue}{0.965}} & \textcolor{blue}{4.078} & \multicolumn{1}{c}{\textcolor{blue}{0.003}} & \multicolumn{1}{c}{\textcolor{blue}{99.73}} & \textcolor{blue}{0.07} &\textcolor{blue}{0.024}\\ 
\bottomrule
\end{tabular}
\end{table*}

\subsection{Ablation Studies and Limitations}\label{sec:ablation}
Below we evaluate the contributions of different components of our architecture, followed by limitation analysis.

\mysubsubsection{Impact of inputs} In Table~\ref{tab:comp}, we obtain our RGB-only or thermal-only variant by removing the corresponding encoders from our architecture. To verify that the performance decrease is not due to the architectural change, we additionally train another two models by inputting the same RGB/thermal images to both encoders (dual-RGB/thermal). Table~\ref{tab:ab_input} shows that the differences between RGB/thermal-only and dual-RGB/thermal are negligible, validating that the accuracy gap is because of using single modality rather than architectural differences.

\begin{figure}[t]
\centering
\includegraphics[width=0.48\textwidth]{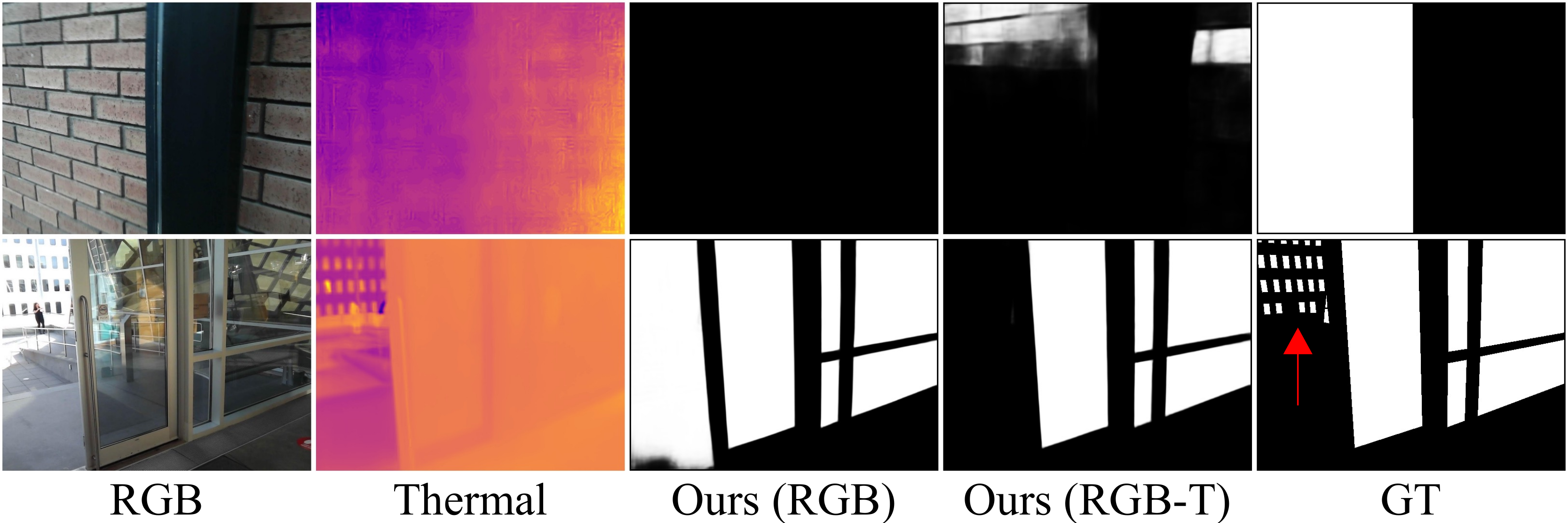}
\caption{Two typical failure examples. The red arrow highlights small glass regions.}
\label{fig:fail}
\end{figure}

\mysubsubsection{Effectiveness of MFM} We compare our method with four simple baselines by replacing the MFM with the simple feature summation (SFS), simple feature concatenation (SFC), the pixel-wise attention fusion (PAF)~\cite{kalra2020deep} and affine transform (AT)~\cite{li2020rgb}. The multi-modal fusion module from EGFNet~\cite{zhou2022edge} for RGB-T fusion are also utilized to replace our MFM. We also modify our MFM by replacing the weighted summation with direct summation (DS) or direct concatenation (DC), where the attention weights are discarded in the two variants. As shown in Table~\ref{tab:ab_mfm}, our variants using the MFM (the last row) achieves the best performance. 

\mysubsubsection{Selection of backbones} Following other related methods~\cite{chen2020dpanet, li2021hierarchical, zhang2020select, zhang2020uc, ji2020accurate}, we exploit a pretrained ResNet backbone. In Table~\ref{tab:ab_backbone}, we evaluate different pretrained backbones and the ResNet-50 backbone without pretraining. We can see that  pretraining is critical for the performance and the ResNet-50 yields the best results. We believe that the performance decrease of ResNet-101 is due to  over-parameterization which makes it hard to train.

\mysubsubsection{Variants of our decoder} Similar to the evaluation of the weighted summation in MFM, we also replace the weighted summation in decoder blocks with direct summation (DS) or direct concatenation (DC). As shown in Table~\ref{tab:ab_decoder}, our decoder that uses weighted summation outperforms the other two variants.

\mysubsubsection{Images without glass} To reduce the false positive segmentation, our training data includes 370 RGB-thermal images pairs without any glass. As shown in Table~\ref{tab:ab_glass}, the model trained on only glass images has an obvious performance decrease on both images with and without glass, which demonstrates the necessity of such samples.

\mysubsubsection{Failure cases} The first row in Fig.~\ref{fig:fail} shows that our method fails when the visual appearances of glass and non-glass regions are highly similar in both RGB and thermal images, which is also a difficult, if not impossible, task for the human eye to differentiate without looking at the GT mask beforehand. Our method also fails to detect small glass regions, as highlighted by the red arrow in the second row.

\section{Application}

\begin{figure*}[!htbp]
\centering
\includegraphics[width=\textwidth]{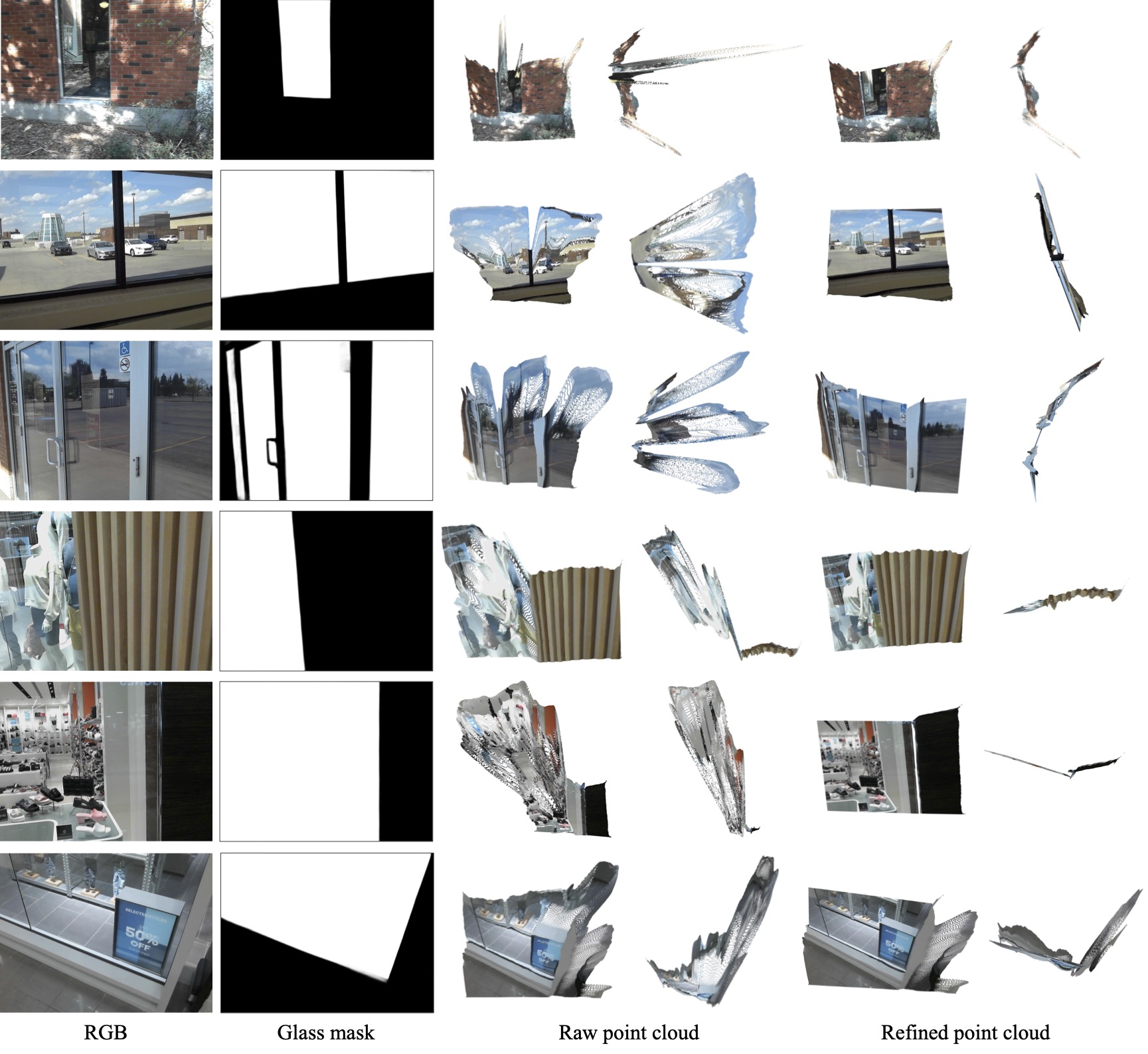}
\caption{Application of monocular 3D reconstruction. From left to right, it shows the RGB images, the glass segmentation masks by our method, the raw reconstructed point clouds by~Adabins~\cite{bhat2021adabins} and our corrected point clouds, respectively.}
\label{fig:depth}
\end{figure*}

\begin{figure*}[!htbp]
\centering
\includegraphics[width=\linewidth]{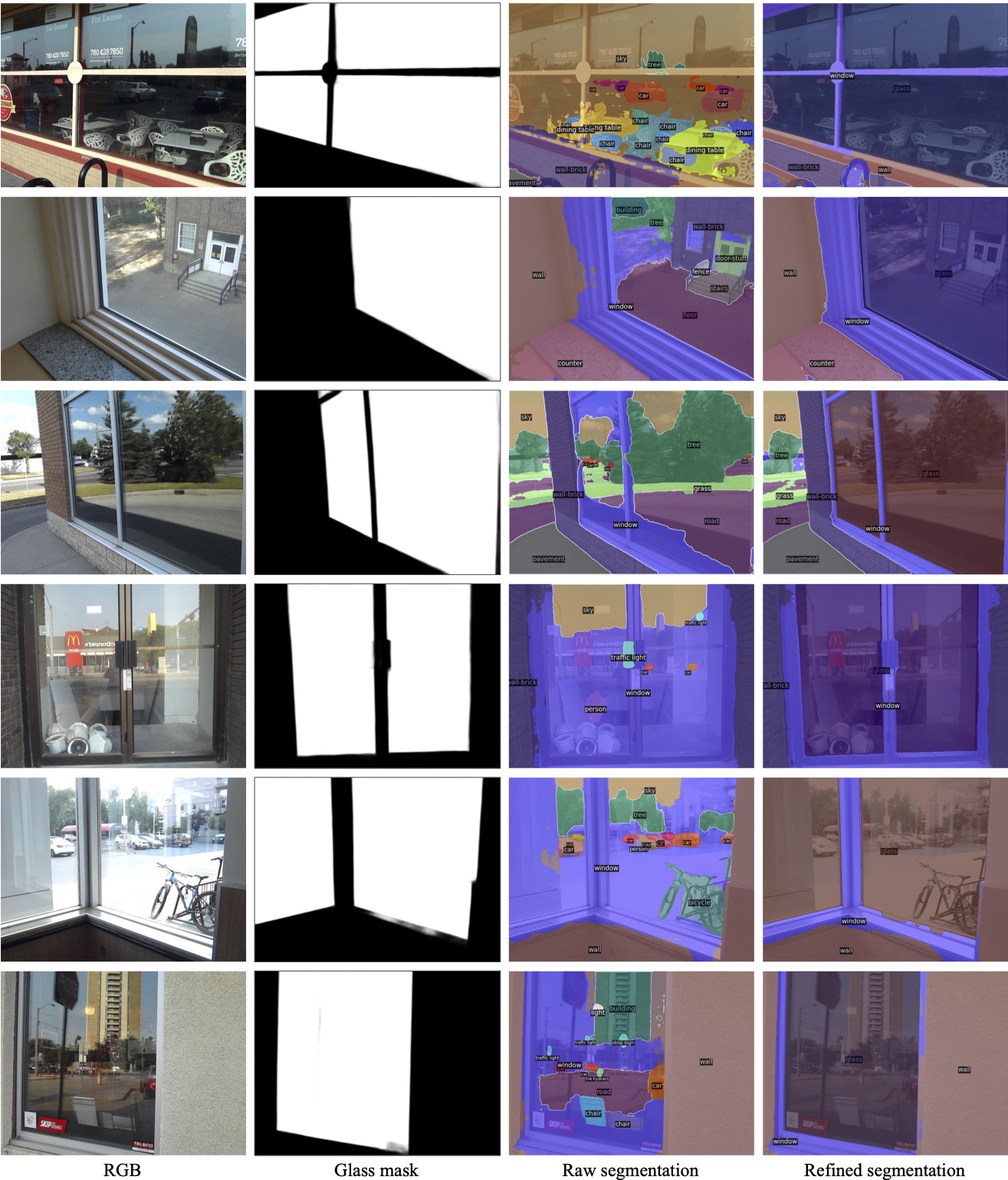}
\caption{Application of semantic image segmentation. From left to right, it shows the RGB images, the glass segmentation masks by our method, the raw semantic segmenation results of DETR~\cite{carion2020end} and our refined results, respectively.}
\label{fig:seg}
\end{figure*}


\subsection{Monocular 3D Reconstruction}
State-of-the-art 3D reconstruction approaches including recent deep-learning-based ones~\cite{bhat2021adabins, ramamonjisoa2021single, zhang2021joint} exhibit challenges when handling scenes with glass (\eg, urban buildings, indoor scenes). While it seems that glass elements occupy a relatively small region in an entire scene, inaccurate glass geometry is catastrophic to the overall 3D reconstruction performance (see Fig.~\ref{fig:depth}), leading to undesirable geometry artifacts and subsequent unpleasant user experiences in downstream applications including augmented reality, gaming, navigation, rendering.

To correct such reconstruction errors in glass regions, we apply our RGB-T glass segmentation method and recover each glass region as a 3D plane. Specifically, we first adopt a recent monocular reconstruction method~\cite{bhat2021adabins} for depth estimation. We then follow Mirror3D~\cite{tan2021mirror3d} and estimate the 3D glass plane parameters based on the depths of the boundary pixels of glass regions. The final 3D point clouds are generated with pre-defined camera intrinsics in Open3D~\cite{zhou2018open3d}. As shown in Fig.~\ref{fig:depth}, by accurately segmenting glass surfaces, our plane-based depth correction strategy significantly improves 3D reconstruction results compared to the raw point clouds from~\cite{bhat2021adabins}.




\subsection{Semantic Segmentation}
Similar problems also arise in semantic image segmentation~\cite{carion2020end, douillard2021plop, zhou2021gmnet}, as shown in Fig.~\ref{fig:seg} where the reflections on glass surfaces are mis-recognized. To correct those errors, we first utilize our RGB-T segmentation method to obtain glass masks and set the glass regions of RGB images to zero, which serve as inputs to a transformer architecture DETR~\cite{carion2020end} (designed for both object detection and semantic segmentation) to get the refined semantic segmentation results. As shown in Fig.~\ref{fig:seg}, such a simple strategy effectively eliminate inaccurate semantic segmentation.
\section{Conclusion and future work}



The paper presents the idea of leveraging RGB-T image pairs for glass segmentation. We propose a novel neural architecture for fusing features of the RGB and thermal modalities. We also contribute the first RGB-T glass scene dataset with GT masks. Our extensive evaluations reveal the significantly better performance of using an RGB-T pair over using a single RGB image, and demonstrate the superiority of our cross-modality fusion method against existing fusion methods using the same RGB-T input. Polarization method has drawbacks in that detection is sparse for a plate glass, which can be confused by glass and other dielectric surfaces, and might not work when transmission $\gg$ reflection. Since polarization and our method use completely different cues, they can be combined as “RGB + polarization + thermal” (polarized RGB like~\cite{kalra2020deep} or polarized thermal~\cite{zhang2018long}) to improve the result, which is an interesting future work. We could also add an invisible thermal light to handle the limitation of our methods, which becomes an active method. As shown in the first row of Fig.~\ref{fig:fail}, glass and non-glass regions are highly similar in both RGB and thermal images. Considering the different smoothness of glass surface and other materials like brick wall, the thermal camera should obtain clear reflection of the thermal light source on glass while a blurry light cone on others.


\section{Acknowledgments}
\noindent The authors would like to thank Steve Sutphen for his technical support and the Natural Sciences and Engineering Research Council of Canada, the University of Alberta and the University of Manitoba for the partial financial funding. 

\bibliographystyle{IEEEtran}
\bibliography{bib}

\vfill

\end{document}